\documentclass{article}
\usepackage[preprint]{neurips_2020}

\usepackage[usenames,dvipsnames]{xcolor}
\definecolor{r0}{RGB}{103,0,31}
\definecolor{r1}{RGB}{178,24,43}
\definecolor{r2}{RGB}{214,96,77}
\definecolor{r3}{RGB}{244,165,130}
\definecolor{r4}{RGB}{253,219,199}
\definecolor{w0}{RGB}{247,247,247}
\definecolor{b4}{RGB}{209,229,240}
\definecolor{b3}{RGB}{146,197,222}
\definecolor{b2}{RGB}{67,147,195}
\definecolor{b1}{RGB}{33,102,172}
\definecolor{b0}{RGB}{5,48,97}
\definecolor{jobcolor}{cmyk}{0,0,0,.95}
\definecolor{joblightcolor}{cmyk}{0,0,0,.95}
\definecolor{abstractcolor}{cmyk}{0,0,0,.95}
\definecolor{copyrightcolor}{cmyk}{.04,.04,.12,.08}

\usepackage[utf8]{inputenc}
\usepackage[inline]{enumitem}
\usepackage{hyperref}
\hypersetup{colorlinks,linkcolor={black},citecolor={b1},urlcolor={black}} 

\usepackage{natbib}
\usepackage{graphicx}
\usepackage{caption}
\usepackage{subcaption}
\usepackage{amsmath}
\usepackage{amsfonts}
\usepackage{multirow}
\usepackage{mathtools}
\usepackage{todonotes}
\usepackage{fancyhdr}
\usepackage{wrapfig}

\usepackage{totcount}
\newtotcounter{citenum}
\def\oldcite{}
\let\oldcite=\bibcite
\def\bibcite{\stepcounter{citenum}\oldcite}
\renewcommand\refname{References (\total{citenum})}

\title{Causal Machine Learning \\ for Healthcare and Precision Medicine}

\author{%
  Pedro Sanchez \\
  University of Edinburgh \\
  \texttt{pedro.sanchez@ed.ac.uk} \\
  \And
   Jeremy P.~Voisey \\
   Canon Medical Research Europe \\
  \And
  Tian Xia \\
  University of Edinburgh \\
  \And
  Hannah I.~Watson \\
  Canon Medical Research Europe \\
  \And
  Alison Q.~ O'Neil \\
  Canon Medical Research Europe \\
  \And
  Sotirios A.~ Tsaftaris \\
  University of Edinburgh \\
  The Alan Turing Institute \\
}

\begin{document}

\maketitle


\begin{abstract}
    Causal machine learning (CML) has experienced increasing popularity in healthcare. Beyond the inherent capabilities of adding domain knowledge into learning systems, CML provides a complete toolset for investigating how a system would react to an intervention (e.g.\ outcome given a treatment). Quantifying effects of interventions allows actionable decisions to be made whilst maintaining robustness in the presence of confounders. Here, we explore how causal inference can be incorporated into different aspects of clinical decision support (CDS) systems by using recent advances in machine learning. Throughout this paper, we use  Alzheimer's disease (AD) to create examples for illustrating how CML can be advantageous in clinical scenarios. Furthermore, we discuss important challenges present in healthcare applications such as processing high-dimensional and unstructured data, generalisation to out-of-distribution samples, and temporal relationships,  that despite the great effort from the research community remain to be solved. Finally, we review lines of research within causal representation learning, causal discovery and causal reasoning which offer the potential towards addressing the aforementioned challenges.
\end{abstract}




\section{Introduction}
\label{sec:intro}

Considerable progress has been made in predictive systems for medical imaging following the advent of powerful machine learning (ML) approaches such as deep learning \citep{litjens2017survey}. In healthcare, clinical decision support (CDS) tools make predictions for tasks such as detection, classification and/or segmentation from electronic health record (EHR) data such as medical images, clinical free-text notes, blood tests, and genetic data. These systems are usually trained with supervised learning techniques. However, most CDS systems powered by ML techniques learn only associations between variables in the data, without distinguishing between causal relationships and (spurious) correlations. 

CDS systems targeted at precision medicine (also known as personalised medicine) need to answer complex queries about how individuals would respond to interventions. A precision CDS system for Alzheimer's disease (AD), for instance, should be able to \emph{quantify} the effect of treating a patient with a given drug on the final outcome, e.g.\ predict the subsequent cognitive test score. Even with the appropriate data and perfect performance, current ML systems would predict the best treatment based only on previous correlations in data, which may not represent \emph{actionable} information. Information is defined as \emph{actionable} when it enables treatment (interventional) decisions to be based on a comparison between different scenarios (e.g.\ outcomes for treated vs not treated) for a given patient. Such systems need causal inference (CI) in order to make actionable and individualised treatment effect predictions \citep{Bica2021}.

A major upstream challenge in healthcare is how to acquire the necessary information to causally reason about treatments and outcomes. Modern healthcare data is multimodal, high-dimensional and often unstructured. Information from medical images, genomics, clinical assessments, and demographics must be taken into account when making predictions. A multimodal approach better emulates how human experts use information to make predictions. In addition, many diseases are progressive over time, thus necessitating that time (the temporal dimension) is taken into account. Finally, any system must ensure that these predictions will be generalisable across deployment environments such as different hospitals, cities, or countries.

Interestingly, it is the connection between causal inference and machine learning that can help alleviate these challenges. ML allows causal models to process high-dimensional and unstructured data by learning complex non-linear relations between variables. CI adds an extra layer of understanding about a system with expert knowledge, which improves information merging from multimodal data, generalisation, and explainability of current ML systems.

\begin{figure}[t]
    \centering
    \includegraphics[width=\linewidth]{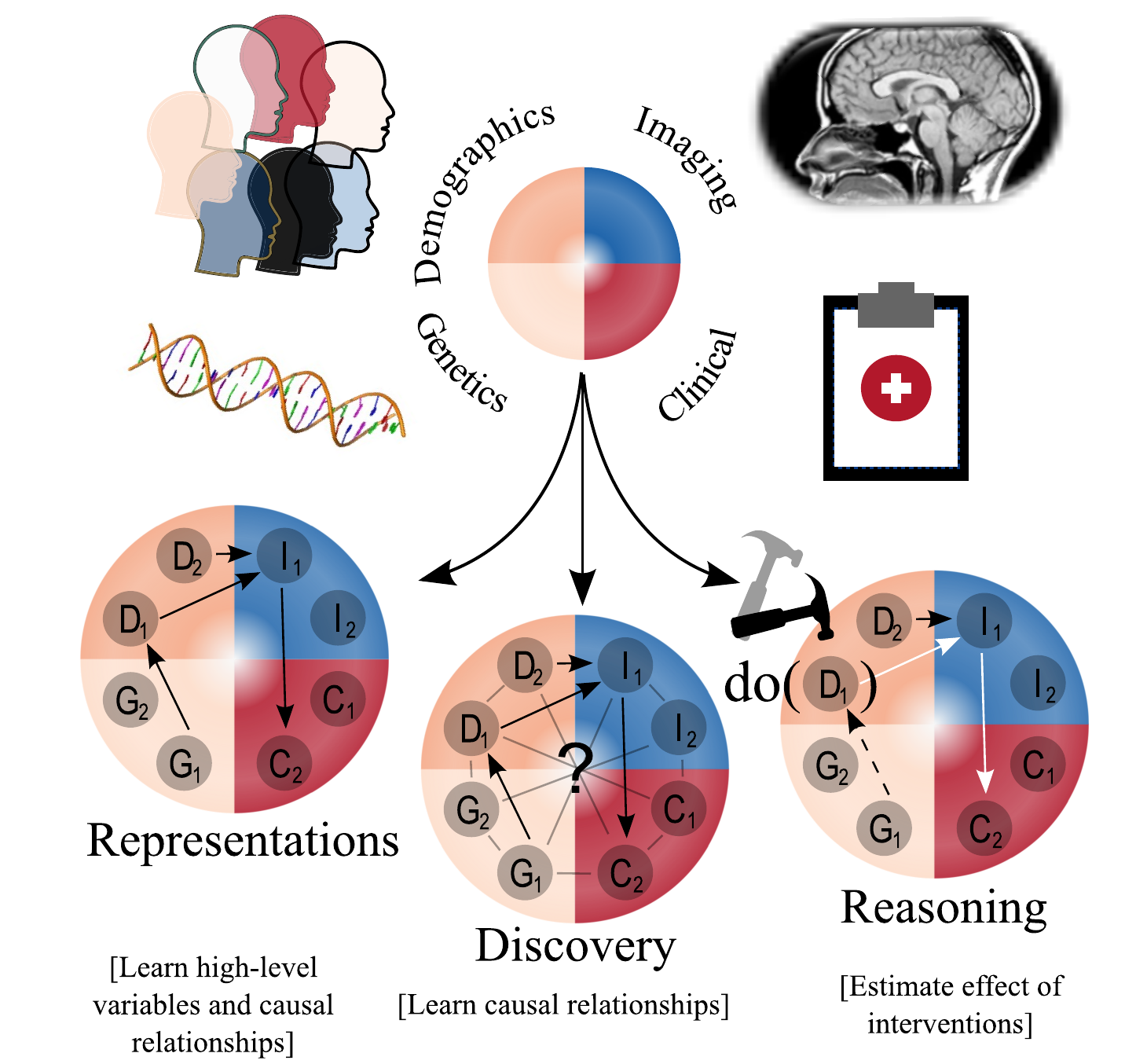}
    \caption{Causal machine learning in healthcare. Healthcare data is multimodal, unstructured and high-dimensional. The research fronts within \textit{causal machine learning} (discussed in Section\ \ref{sec:solutions}) can be further divided into causal representation learning, causal discovery, and causal reasoning.}
    \label{fig:taxonomy}
\end{figure}

The \emph{causal machine learning} (CML) literature offers several directions for addressing the aforementioned challenges when using observational data. Here, we categorise CML into three directions:
\begin{enumerate*}[label=(\roman*)]
    \item \textit{Causal Representation Learning} -- given high-dimensional data, learn to extract low-dimensional informative (causal) variables and their causal relations;
    \item \textit{Causal Discovery} -- given a set of variables, learn the causal relationships between them; and
    \item \textit{Causal Reasoning} -- given a set of variables and their causal relationships, analyse how a system will react to interventions.
\end{enumerate*}
These directions are illustrated in Fig.\ \ref{fig:taxonomy}.

In this paper, we discuss how CML can improve personalised decision making as well as help to mitigate pressing challenges in clinical decision support systems. We review representative methods for CML, explaining how they can be used in a healthcare context. In particular, we 
\begin{enumerate*}[label=(\roman*)]
    \item present the concept of causality and causal models;
    \item show how they can be useful in healthcare settings;
    \item discuss pressing challenges; and
    \item review potential research directions from CML.
\end{enumerate*}

\section{What is causality?}
\label{sec:whatscausality}
We use a broad definition of causality: if $A$ is a cause and $B$ is an effect, then $B$ relies on $A$ for its value. As causal relations are directional, the reverse is not true; $A$ does not rely on $B$ for its value. The notion of \emph{causality} thus enables analysis of how a system would respond to an \emph{intervention}.

Questions such as ``How will this disease progress if a patient is given treatment X?'' or ``Would this patient still have experienced outcome Z if treatment Y was received?'' require methods from causality to understand how an intervention would affect a specific individual. In a clinical environment, causal reasoning can be useful for deciding which treatment will result in the best outcome. For instance, in an AD scenario, causality can answer queries such as ``Which of drug $A$ or drug $B$ would best minimise the patient's expected cognitive decline within a 5 year time span?''.
Ideally, we would compare the outcomes of alternative treatments using observational (historical) data. However, the ``fundamental problem of causal inference'' \citep{10.2307/2289064} is that for each unit (i.e.\ patient) we can observe either the result of treatment $A$ or of treatment $B$, but never both at the same time. This is because after making a choice on a treatment, we cannot turn back time to undo the treatment. These queries that entertain hypothetical scenarios about individuals are called \emph{potential outcomes}. Thus, we can observe only one of the potential consequences of an action; the unobserved quantity becomes a \textit{counterfactual}. Causality's mathematical formalism pioneered by Judea Pearl \citep{pearl_2009} and Donald Rubin \citep{Imbens2015Causal} allows these more challenging queries to be answered.

Most machine learning approaches are not (currently) able to identify cause and effect, because causal inference is fundamentally impossible to achieve without making assumptions \citep{pearl_2009,peters2017elements}. Several of these assumptions can be satisfied through study design or external contextual knowledge, but none can be discovered solely from observational data.

Next, we introduce the reader to two ways of defining and reasoning about causal relationships: with structural causal models and with potential outcomes. We wrap up this section with an introduction to determining causal relationships, including the use of randomised controlled trials.

\subsection{Structural Causal Models}
\label{subsec:scms}

The mathematical formalism around the so-called \textit{do-calculus} and \emph{structural causal models} (SCMs) pioneered by the Turing Award winner Judea Pearl \citep{pearl_2009} has allowed a graphical perspective to reasoning with data which heavily relies on domain knowledge. This formalism can model the data generation process and incorporate assumptions about a given problem. An intuitive and historical description of causality can be found in \citet{Pearl_why}'s recent book \emph{The Book of Why}.

An SCM $\mathfrak{G} := (\mathbf{S},P_{\boldsymbol{N}})$ consists of a collection $\mathbf{S} = (f_1,....,f_K)$  of structural assignments (called mechanisms)
\begin{equation}
    X_k := f_k(\mathbf{PA}_k,N_k),
\end{equation}
where $\mathbf{PA}_k$ is the set of parent variables of $X_k$ (its direct causes) and $N_k$ is a noise variable for modeling uncertainty. $\boldsymbol{N} = \{N_1,N_2,...,N_d\}$ is also referred to as \emph{exogenous} noise because it represents variables that were not included in the causal model, as opposed to the \emph{endogenous} variables $\boldsymbol{X} = \{X_1,X_2,...,X_d\}$ which are considered known or at least intended by design to be considered, and from which the set of parents $Pa_k$ are drawn. This model can be defined as a direct acyclic graph (DAG) in which the nodes are the variables and the edges are the causal mechanisms. One might consider other graphical structures which incorporate cycles and latent variables \citep{Bongers2021cycles}, depending on the nature of the data.

It is important to note that the causal mechanisms are representations of physical mechanisms that are present in the real world. Therefore, according to the principle of \emph{independent causal mechanisms} (ICM), we assume that the causal generative process of a system’s variables is composed of autonomous modules that do not inform or influence each other \citep{peters2017elements,scholkopf2021toward}. This means that exogenous variables $\boldsymbol{N}$ are mutually independent with the following joint distribution $P(\boldsymbol{N}) = \prod_{k=1}^{d} P(N_k)$. Moreover, the joint distribution over the endogenous variables $\boldsymbol{X}$ can be factorised as a product of independent conditional mechanisms
\begin{equation}
    P_{\mathfrak{G}}(X_1,X_2,..,X_K)=\prod_{k=1}^{K} P_{\mathfrak{G}}\left(X_{k} \mid \mathbf{PA}_{k}\right) .
\end{equation}

The causal framework now allows us to go beyond
\begin{enumerate*}[label=(\roman*)]
    \item associative predictions, and begin to answer
    \item interventional and
    \item counterfactual
\end{enumerate*}
queries. These three tasks are also known as Pearl's \textit{causal hierarchy} \citep{Pearl_why}. The \emph{do-calculus} introduces the notation $do(A)$, to denote a system where we have \emph{intervened} to fix the value of $A$. This allows us to sample from an interventional distribution $P^{\mathfrak{G};do(...)}_{\boldsymbol{X}}$, which has the advantage over an observational distribution $P^{\mathfrak{G}}_{\boldsymbol{X}}$ that the causal structure enforces that only the descendants of the variable intervened upon will be modified by a given action. 

\subsection{Potential Outcomes}
\label{subsec:potential_outcomes}
An alternative approach to causal inference is the \emph{potential outcomes} framework proposed by \citet{rubin2005causal}. In this framework, a response variable $Y$ is used to measure the effect of some cause or treatment for a patient, $i$. The value of $Y$ may be affected by the treatment assigned to $i$. To enable the treatment effect to be modelled, we represent the response with \emph{two} variables $Y_{i}^{(0)}$ and $Y_{i}^{(1)}$ which denote ``untreated'' and ``treated'' respectively. The effect of the treatment on $i$ is then the difference, $Y_{i}^{(1)} - Y_{i}^{(0)}$.
     
As a patient may \emph{potentially} be untreated or treated, we refer to $Y_{i}^{(0)}$ and $Y_{i}^{(1)}$ as \emph{potential outcomes}. It is, however, impossible to observe both simultaneously, according to the previously mentioned \emph{fundamental problem of causal inference} \citep{10.2307/2289064}. This does not mean that causal inference itself is impossible, but it does bring challenges \citep{Imbens2015Causal}. Causal reasoning in the potential outcome frameworks depends on obtaining an estimate for the joint probability distribution, $P(Y^{(0)}, Y^{(1)})$.

Both SCM and potential outcomes approaches have useful applications, and are used where appropriate throughout this article. We note that single world intervention graphs \citep{richardson2013single} have been proposed as a way to unify them.

\subsection{Determining Cause and Effect} 
\label{subsec:randomised_control_trial}

Determining causal relationships often requires carefully designed experiments. There is a limit to how much can be learned using purely observational data.

The effects of causes can be determined through prospective experiments to observe an effect $E$ after a cause $C$ is tried or withheld, keeping constant all other possible factors. It is hard, and in most cases impossible, to control for all possible confounders of $C$ and $E$. The gold standard for discovering a true causal effect is by performing a randomised controlled trial (RCT), where the choice of $C$ is randomised, thus removing confounding. For example, by randomly assigning a drug or a placebo to patients participating in an interventional study, we can measure the effect of the treatment, eliminating any bias that may have arisen in an observational study due to other confounding variables, such as lifestyle factors, that influence both the choice of using the drug and the impact of cognitive decline \citep{Mangialasche2010}.

Note that the conditional probability $P(E \mid C)$ of observing $E$ after observing $C$ can be different from the interventional probability $P(E \mid do(C))$ of observing $E$ after doing / intervening on $C$. $P(E \mid do(C))$ means that only the descendants of $C$ (in a causal graph) change after an intervention, all other variables maintain their values. In RCTs, the ‘$do$’ is guaranteed and unconditioned, while with observational data such as historical Electronic Health Records (EHRs), it is not, due to the presence of confounders.

Determining the causes of effects (the aetiology of diseases) requires hypotheses and experimentation where interventions are performed and studied to determine the necessary and sufficient conditions for an effect or disease to occur.

\section{Why should we consider a causal framework in healthcare?}
\label{sec:causal_healthcare}

Causal inference has made several contributions over the last few decades to fields such as social sciences, econometrics, epidemiology, and aetiology  \citep{pearl_2009,Imbens2015Causal}, and it has recently spread to other healthcare fields such as medical imaging \citep{castro2020causality,Pawlowski,Reinhold2021} and pharmacology \citep{Bica2021}. In this section, we will elaborate on how causality can be used for improving medical decision making. 

Even though data from EHRs, for example, are usually observational, they have already been successfully leveraged in several machine learning applications \citep{PICCIALLI2021111}, such as modeling disease progression \citep{Lim2018}, predicting disease deterioration \citep{prediction_kidney_nature}, and discovering risk factors \citep{McCauley2016}, as well as for predicting treatment responses \citep{Athreya_1482}. Further, we now have evidence of algorithms which achieve superhuman performance in imaging tasks such as segmentation \citep{isensee2021nnu}, detection of pathologies and classification \citep{korot2021code}. However, predicting a disease with almost perfect accuracy for a given patient is not what precision medicine is trying to achieve \citep{Wilkinson2020}. Rather, we aim to build machine learning methods which extract \emph{actionable} information from observational patient data in order to make interventional (treatment) decisions. This requires \emph{causal inference}, which goes beyond standard supervised learning methods for prediction as detailed below.

In order to make actionable decisions at the patient level, one needs to estimate the treatment effect. The treatment effect is the \emph{difference} between two potential outcomes: the \emph{factual} outcome and the \emph{counterfactual} outcome. For actionable predictions, we need algorithms that learn how to reason about hypothetical scenarios in which different actions could have been taken, creating, therefore, a decision boundary that can be navigated in order to improve patient outcome. There is recent evidence that humans use counterfactual reasoning to make causal judgements \citep{Gerstenberg2021csm}, lending support to this reasoning hypothesis.

This is what makes the problem of inferring treatment effect fundamentally different from standard supervised learning \citep{Bica2021} as defined by the potential outcome framework \citep{rubin2005causal,Imbens2015Causal}. When using observational datasets, by definition, we never observe the counterfactual outcome. Therefore, the best treatment for an individual -- the main goal of precision medicine \citep{zhang2018learning} -- can only be identified with a model that is capable of causal reasoning as will be detailed in Section\ \ref{subsec:precision_medicine}.

\subsection{Alzheimer's Disease practical example} 

We now illustrate the notion of causal machine learning for healthcare with an example from \textit{Alzheimer's disease} (AD). A recent attempt to understand AD from a causal perspective \citep{Shen2020Discovery,Uleman2020Alzheimer} takes into account many biomarkers and uses domain knowledge (as opposed to RCTs) for deriving ground truth causal relationships. In this section, we present a simpler view with only three variables: chronological age\footnote{Age can otherwise be measured in biological terms using, for instance, DNA methylation \citep{Horvath2013DNAm}.}, magnetic resonance (MR) images of the brain, and Alzheimer's disease diagnosis. The diagnosis of Alzheimer's disease is made by a clinician who takes into account all available clinical information, including images. We are particularly interested in MR images because analysing the relationship of high-dimensional data such as medical images, is a task that can be more easily handled with machine learning techniques, the main focus of this paper.

\begin{figure}[t]
    \centering
    \includegraphics[width=\linewidth]{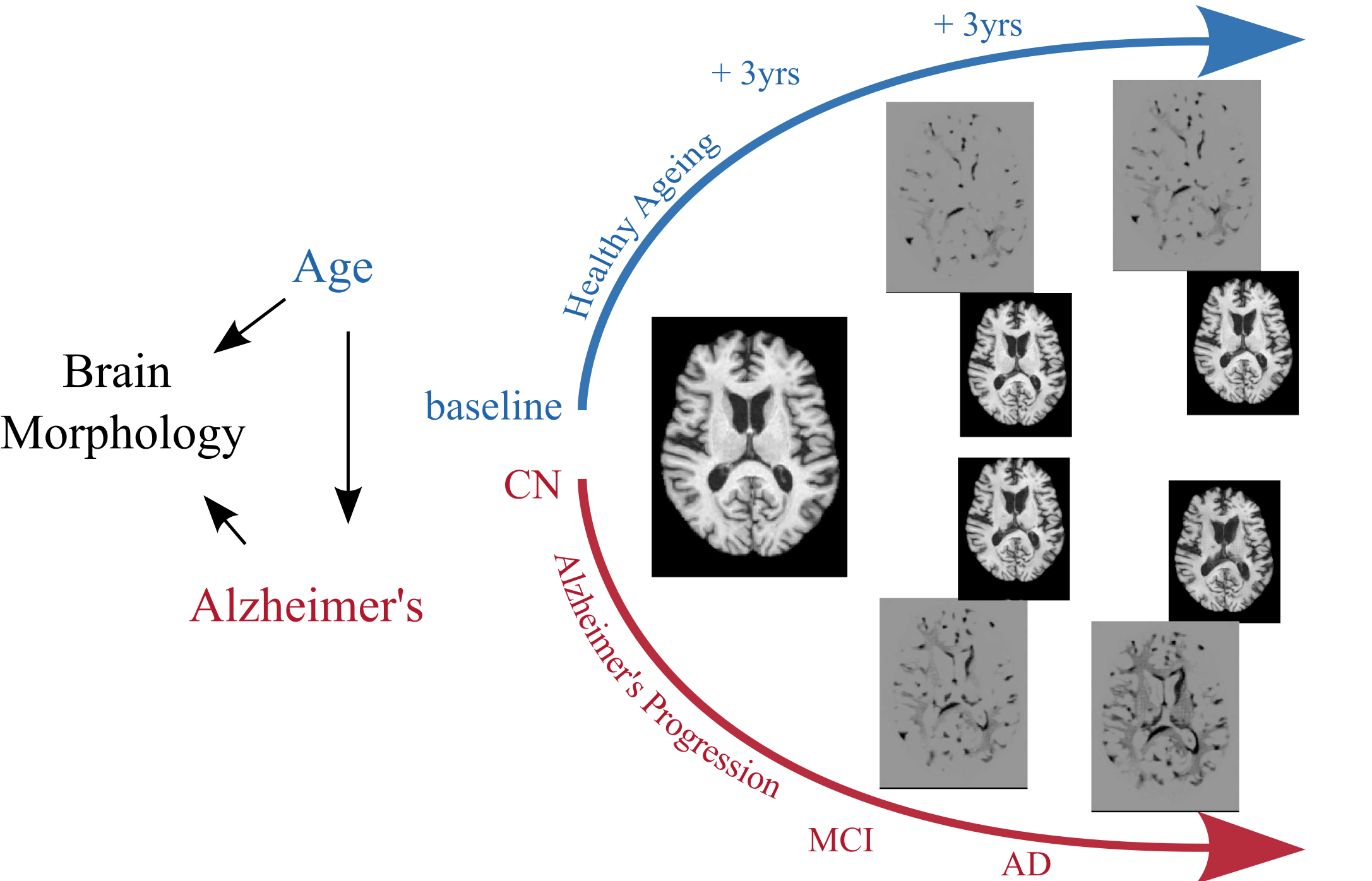}
    \caption{Causal graph (\textit{left}) and illustration of how the brain changes in MR images in response to interventions on ``Age" or ``Alzheimer's disease status". The images are axial slices of a brain MR scan. The middle image used as a baseline is from a patient aged 64 years old who is classified an cognitively normal (CN) within the Alzheimer’s Disease Neuroimaging Initiative (ADNI) database. All other images are synthesised with a conditional generative model \citep{xia2021learning}. The images with gray background are difference images obtained by subtracting the synthesised image from the baseline. The upper sequence of images is generated by fixing Alzheimer's status at CN and increasing age by 3 years. The bottom images are generated by fixing the age at 64 and increasing Alzheimer's status to MCI and AD, as discussed in the main text.}
    \label{fig:alzheimers_scm}
\end{figure}

AD is a type of cognitive decline that generally appears later in life \citep{Jack2015JAMA}. Alzheimer's disease is associated with brain atrophy \citep{Karas2004NeuroImage,Qian2019} i.e.\ volumetric reduction of grey matter. We consider that Alzheimer's disease causes the symptom of brain morphology change, following \citet{Richens2021nature}, by arguing that a high-dimensional variable such as the MR image is caused by the factors that generated it; this modelling choice has been previously used in the causality literature \citep{Scholkopf2012anticausal, kilbertus2018generalization,Deml2021Anti}. Further, it is well established that atrophy also occurs during normal ageing \citep{Sullivan1995Neurobiology,Good2001Morphometric}. Time does not depend on any biological variable, therefore chronological age cannot be caused by Alzheimer's disease nor any change in brain morphology. In this scenario, we can assume that age is a confounder of brain morphology, measured by the MR image, and Alzheimer's disease diagnosis. These relationships are illustrated in the causal graph in Fig.\ \ref{fig:alzheimers_scm}.

To model the effect of having age as a confounder of brain morphology and Alzheimer's disease, we use a conditional generative model from \citet{xia2021learning}\footnote{We take the model from \citet{xia2021learning} and run new demonstrative experiments for illustration in this paper.}, in which we condition on age and Alzheimer's disease diagnosis for brain MRI image generation. We then synthesise images of a patient at different ages and with different Alzheimer's disease status as depicted in Fig.\ \ref{fig:alzheimers_scm}. In particular, we control for (i.e.\ condition on) one variable while intervening on the other. That is, we synthesise images based on a patient who is cognitively normal (CN) for their age of 64 years. We then fix the Alzheimer's status at CN and increase the age by 3 years for 3 steps, resulting in images of the same CN patient at ages 64, 67, 70, 73. At the same time, we synthesise images with different Alzheimer's status by fixing the age at 64 and changing the Alzheimer's status from mild cognitive impairment (MCI) to a clinical diagnosis of Alzheimer's disease (AD).

This example illustrates the effect of \textit{confounding bias}. By observing qualitatively the difference between the baseline and synthesised images, we see that ageing and Alzheimer's disease have similar effects on the brain\footnote{See \citet{xia2021learning} for quantitative results confirming this hypothesis.}. That is, that both variables change the volume of brain when intervened on independently. 

Throughout the paper, we will further add variables and causal links to this example to illustrate how healthcare problems can become more complex and how a causal approach might mitigate some of the main challenges. In particular, we will build on this example by explaining some consequences of causal modelling for dealing with high-dimensional and unstructured data, generalisation and temporal information.

\subsection{Modeling the Data Generation Process}
\label{subsec:data_gen_process}

The Alzheimer's disease example illustrates the importance of considering causal relationships in a machine learning scenario. Namely, causality gives the ability to model and identify types and sources of bias\footnote{we refer to \url{https://catalogofbias.org/biases} for a catalogue of bias types}. To correctly identify which variables to control for (as means to mitigate confounding bias), causal diagrams \citep{pearl_2009} offer a direct means of visual exploration and consequently explanation \citep{Brookhart2010,Lederer2019Annals}.

\citet{castro2020causality} details further how understanding the causal generating process can be useful in medical imaging. By representing the variables of a particular problem and their causal relationships as a causal graph, one can model \emph{domain shifts} such as population shift (different cohorts), acquisition shift (different sites or scanners) and annotation shift (different annotators), and data scarcity (imbalanced classes).

In the Alzheimer's disease setting above, a classifier naively trained to perform diagnosis from MR images of the brain might focus on the brain atrophy alone.
This classifier may show reduced performance in younger adults with Alzheimer's disease or for cognitively normal older adults, leading to potentially incorrect diagnosis. To illustrate this, we report the results of a convolutional neural network (CNN) classifier trained and tested on the ADNI dataset following the same setting as \citet{xia2022adversarial}\footnote{Although we replicate results from \citet{xia2022adversarial}, this work does not constitute an extension of the original paper. Rather, we use \citet{xia2022adversarial} as an example that illustrates how causality might impact standard machine learning.}.
Table\ \ref{tab:alzheimer_class} shows that as feared, healthy older patients (80-90 years old) are less accurately predicted because ageing itself causes the brain to have Alzheimer's-like patterns.

\begin{table}[t]
\caption{Illustration of how a naively trained classifier (a neural network) fails when the data generation process and causal structure are not identified. We report the precision and recall on the test set when training a classifier for diagnosing AD. We stratify the results by age. We highlight that the group with {\color{r2} worse performance} is the older cognitively normal patients due to the confounding bias described in the main text. After training with counterfactually augmented data, the classifier's precision for the worse performance age group improved. These results were replicated from \citet{xia2022adversarial}.}

\label{tab:alzheimer_class}
\begin{center}
\begin{tabular}{cclll}
\hline
\multicolumn{2}{c}{\textbf{Age Range (years)}} & 60-70 & 70-80 & 80-90                       \\ \hline
                           & Precision & 87.7  & 91.4  & {\color{r2} 75.5} \\ \cline{2-5} 
\multirow{-2}{*}{Naive} &  Recall & 92.5  & 94.2  & 97.1 \\ \hline
 & Precision & 88.3  & 93.6  & {\color{b2} 84.2} \\ \cline{2-5} 
\multirow{-2}{*}{Counterfactually Augmented} &  Recall & 91.5  & 96.5  & 95.7 \\ \hline

\end{tabular}
\end{center}
\end{table}

Indeed, using augmented data based on causal knowledge is a solution discussed in \citet{xia2022adversarial}, whereby the training data are augmented with counterfactual images of a patient when intervening on age. That is, images of a patient at different ages (while controlling for Alzheimer's status) are synthesised so the classifier learns how to differentiate the effects of ageing vs Alzheimer's disease in brain images.

This causal knowledge enables the formulation of best strategies for mitigating data bias(es) and improving generalisation (further detailed in Section\ \ref{subsec:generalisation}). For example, if after modeling the data distribution, an acquisition shift becomes apparent (e.g.\ training data were obtained with a specific MR sequence but the model will be evaluated on data from a different sequence), then data augmentation strategies can be designed to increase robustness of the learned representation. The acquisition shift -- e.g.\ different intensities due to different scanners -- might be modeled according to the physics of the (sensing) systems. Ultimately, creating a diagram of the data generation process helps rationalise/visualise which are the best strategies to solve the problem.

\subsection{Treatment Effect and Precision Medicine}
\label{subsec:precision_medicine}

\begin{figure}[h]
    \centering
    \includegraphics[width=0.9\linewidth]{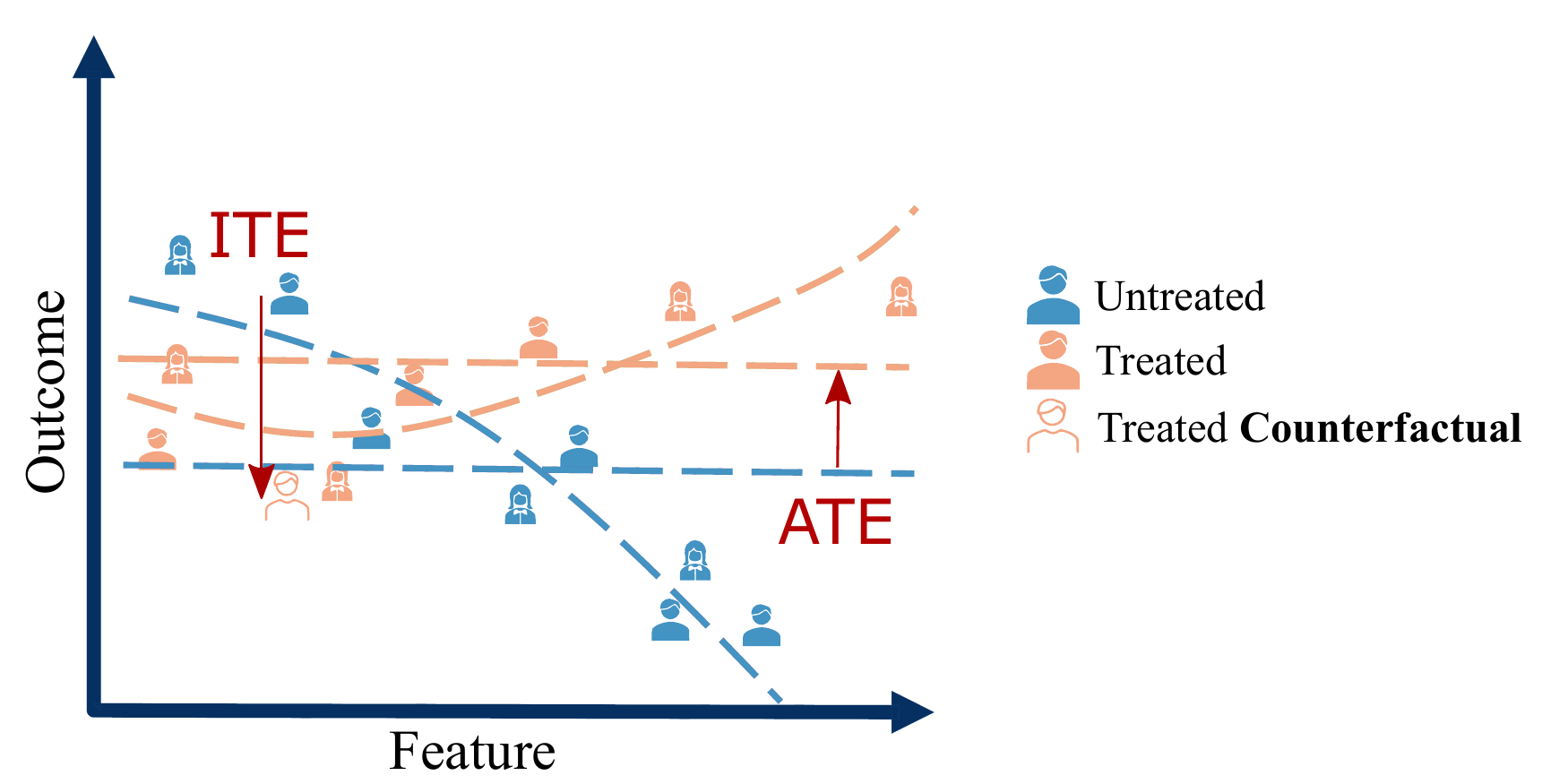}
    \caption{We illustrate the difference between individualised and average treatment effect (ITE vs ATE). ``Feature'' represents patient characteristics, which would be multi-dimensional in reality. ``Outcome'' is some measure of response to the treatment, where a more positive value is preferable. The ITE for each patient is the difference between actual and the counterfactual outcome. We show an example counterfactual to highlight that ITE for some patients might differ from the average (ATE). By employing causal inference methods to estimate individualised treatment effects, we can understand which patients benefit from certain medication and which patients do not, thus enabling us to make personalised treatment recommendations. Note that the patient data points are evenly distributed along the feature axis which would indicate that this data comes from an RCT (due to lack of bias). The estimation of treatment affect using observational data is subject to confounding as patient characteristics effect both the selection of treatment and outcome. Causal inference methods need to mitigate this.
    }
    \label{fig:ite}
\end{figure}

Beyond diagnosis, a major challenge in healthcare is ascertaining whether a given treatment influences an outcome. For a binary treatment decision, for instance, the aim is to estimate the \emph{average treatment effect} (ATE), $E[Y^{(1)} - Y^{(0)}]$ where $Y^{(1)}$ is the outcome given the treatment and $Y^{(0)}$ is the outcome without it (control). As it is impossible to observe both potential outcomes $Y^{(0)}$ and $Y_{i}^{(1)}$ for a given patient $i$, this is typically estimated using $E[Y|T=1] - E[Y|T=0]$, where $T$ is the treatment assignment. 

The treatment assignment and outcomes, however, both depend on the patient's condition in normal clinical conditions. This results in confounding, which is best mitigated by the use of an RCT \ref{subsec:randomised_control_trial}. Performing an RCT as detailed in Section\ \ref{subsec:randomised_control_trial}, however, is not always feasible, and causal inference techniques can be used to estimate the causal effect of treatment from observational data \citep{Meid2020}. A number of assumptions need to hold in order for the treatment effect to be identifiable from observational data \citep{Lesko2017, Imbens2015Causal}. Conditional exchangeability (ignorability) assumes there are no unmeasured confounders. Positivity (overlap) is the assumption that every patient has a chance of receiving each treatment. Consistency assumes that the treatment is defined unambiguously.  Continuing the Alzheimer's example, \citet{Charpignon2021} explore drug re-purposing by emulating an RCT with a target trial \citep{hernan2016using} and find indications that metformin (a drug classically used for diabetes) might prevent dementia.

Note that even if the treatment effect is estimated using data from a well-designed RCT, $ E[Y \mid T=1] - E[Y \mid T=0] $ is the \emph{average} treatment effect across the study population. However, there is evidence \citep{Bica2021} that for any given treatment, it is likely that only a small proportion of subjects will actually respond in a manner that resembles the ``average'' patient, as illustrated in Fig.\ \ref{fig:ite}. In other words, the treatment effect can be highly heterogeneous across a population. The aim of \emph{precision medicine} is to determine the best treatment for an \emph{individual} \citep{Wilkinson2020}, rather than simply measuring the average response across a population. In order to answer this question for a binary treatment decision, it is necessary to estimate $ \tau_i = Y_i(1) - Y_i(0)$ for a patient $i$. This is known as the individualised treatment effect (ITE). As this estimation is performed using a conditional average, this is also referred to as the conditional average treatment effect (CATE) \citep{abrevaya2015estimating}. 

 
A number of approaches have been proposed to learn conditional average treatment effect from observational data, such as estimating treatment effect with double machine learning \citep{Chernozhukov2018-mt,Semenova2020}. Another trend for estimating CATE are based on meta-learners \citep{kunzel2019metalearners,curth2021nonparametric}. In the meta-learning setting, traditional (supervised) machine learning is used to predict the conditional expectations of outcome for units under control and under treatment separately. Then, CATE is done by taking the difference between the estimates of these estimates. While most approaches concentrate on estimating CATE using observational data, it is also possible to do so using data from an RCT \citep{Hoogland2021}.

A long-term goal of precision medicine \citep{Bica2021} includes personalised risk assessment and prevention. Without a causal model to distinguish these questions from simpler prediction systems, interpretational mistakes will arise. In order to design more robust and effective machine learning methods for personalised treatment recommendations, it is vital that we gain a deeper theoretical understanding of the challenges and limitations of modeling multiple treatment options, combinations, and treatment dosages from observational data.

\section{Causal machine learning for complex data}
\label{sec:complex_data}



In Section \ref{sec:causal_healthcare}, we focused on causal reasoning in situations where the causal models are known (at least partially) and variables are well demarcated. We refer the reader to \citet{Bica2021} for a comprehensive review on these methods. Most healthcare problems, however, have challenges that are upstream of causal reasoning.
In this section, we highlight the need to deal with high-dimensional and multimodal data as well as with temporal information and discuss generalisation in out-of-distribution settings when learning from unstructured data.

\subsection{Multimodal Data}
\label{subsec:dimension_modal}

Alzheimer's disease, in common with other major diseases such as diabetes and cancer, has multiple causes arising from complex interactions between genetic and environmental factors. Indeed, a recent attempt \citep{Shen2020Discovery} to build causal graphs for describing Alzheimer's disease takes into account data derived from several data sources and modalities, including patient demographics, clinical measurements, genetic data, and imaging exams. \citet{Uleman2020Alzheimer}, in particular, creates a causal graph \footnote{Interestingly, \citet{Uleman2020Alzheimer} gather expert knowledge using a group model-building technique \citep{Vennix1999} where multiple experts with complementary skills create a graph based on their combined mental models and assumptions.} with clusters of nodes related to brain health, physical health, and psychosocial health, illustrating the complexity of AD.


The above example illustrates that modern healthcare is multimodal. New ways of measuring biomarkers are increasingly accessible and affordable, but integrating this information is not trivial. Information from different sources needs to be transformed to a space where information can be combined, and the common information across modalities needs to be disentangled from the unique information within each modality \citep{braman2021deep}. This is critical for developing CDS systems capable of integrating images, text and genomics data.

On the other hand, the availability of more variables might mean that some assumptions which are made in classical causal inference are more realistic. In particular, most methods consider the assumption of \emph{conditional exchangeability} (or causal \emph{sufficiency} \citep{spirtes2000causation}), as in section \ref{sec:causal_healthcare} \ref{subsec:precision_medicine}. In practice, the conditional exchangeability assumption may often not be true due to the presence of unmeasured confounders. However, observing more variables might reduce the probability of this, rendering the assumption more plausible.

\subsection{Temporal Data}

It is well known that a gene called apolipoprotein E (APOE) is associated with an increased risk of AD \citep{Zlokovic2013,Mishra2018}. However, environmental factors, such as education \citep{Stern1994,Larsson2017,Anderson2020}, also have an impact on dementia. In other words, environmental factors over time contribute to different disease trajectories in Alzheimer's disease. In addition, there are possible loops in the causal diagram \citep{Uleman2020Alzheimer}. \citet{Wang2019} illustrate, for instance, a positive feedback loop between sleep and AD. That is, poor sleep quality aggravates amyloid-beta and tau pathology concentrations, potentially leading to neuronal dysfunction which, in turn, leads to worse sleep quality. It is, therefore, important to consider data-driven approaches for understanding and modeling the progression of disease over time \citep{Oxtoby2017neurodegenerative}.

At the same time, using temporal information for inferring causation can be traced back to one of the first definitions of causality by \citet{Hume1904HUMECH}. Quoting \citeauthor{Hume1904HUMECH}: ``we may define a cause to be an object followed by another, and where all the objects, similar to the first, are followed by objects similar to the second''. There are many strategies for incorporating time into causal models since using SCMs with directed acyclic graphs (as defined in Section \ref{sec:whatscausality} \ref{subsec:scms}) is not enough in this context. A classical model of causality for time-series was developed by \citet{granger1969investigating}. \citeauthor{granger1969investigating} considers $X \rightarrow Y$ if past $X$ is predictive of future $Y$. Therefore, inferring causality from time series data is at the core of CML. \citet{Bongers2021cycles} shows that SCMs can be defined with latent variables and cycles, allowing temporal relationships. Early work has used temporal causal inference in neuroscience \citep{friston2003dynamic}, but the application of temporal causal inference in combination with machine learning for understanding and dealing with complex disease remains largely unexplored.

Managing diseases such as Alzheimer's disease can be challenging due to the heterogeneity of symptoms and their trajectory over time across the population. A pathology might evolve differently for patients with different covariates. For treatment decisions in a longitudinal setting, causal inference methods need to model patient history and treatment timing \citep{soleimani2017treatmentresponse}. Estimating trajectories under different possible future treatment plans (interventions) is extremely important \citep{Bica2020Estimating}. CDS systems need to take into account the current health state of the patient, to make predictions about the potential outcomes for hypothetical future treatment plans, to enable decision-makers to choose the sequence and timing of treatments that will lead to the best patient outcome \citep{Lim2018Forecasting,Bica2020Estimating,Rui2021ml4h}. 



\subsection{Out-of-Distribution Generalisation with Unstructured and High-Dimensional Data}
\label{subsec:generalisation}

The challenge of integrating different modalities and temporal information increases when unstructured data is used. Most causality theory was originally developed in the context of epidemiology, econometrics, social sciences, and other fields wherein the variables of interest tend to be scalars \citep{pearl_2009,Imbens2015Causal}. In healthcare, however, the use of imaging exams and free-text reports poses significant challenges for consistent and robust extraction of meaningful information. The processing of unstructured data is mostly tackled with machine learning, and \emph{generalisation} is one of the biggest challenges for learning algorithms.

In its most basic form, generalisation is the ability to correctly categorise new samples that differ from those used for training \citep{bishop2006pattern}. However, when learning from data, the notion of generalisation has many facets. Here, we are interested in a realistic setting where the test data distribution might be different from the training data distribution. This setting is often referred as \textit{out-of-distribution (OOD) generalisation}. Distribution shifts are often caused by a change in environment (e.g.\ different hospitals). We wish to present a causal perspective \citep{Gong2016,Rojas2018jmlr,Meinshausen2018} on generalisation which unifies many machine learning settings. Causal relationships are stable across different environments \citep{Cui2022}. In a causal learning, the prediction should be invariant to distribution shifts \citep{peters2016causal}.

As the use of machine learning in high impact domains becomes widespread, the importance of evaluating safety has increased. A key aspect is evaluating how robust a model is to changes in environment (or domain), which typically requires applying the model to multiple independent datasets \citep{pmlr-v130-subbaswamy21a}. Since the cost of collecting such datasets is often prohibitive, causal inference argues that providing structure (which comes from expert knowledge) is essential for increasing robustness in real life \citep{pearl_2009}.

\begin{figure}[t]
    \centering
    \includegraphics[width=.5\linewidth]{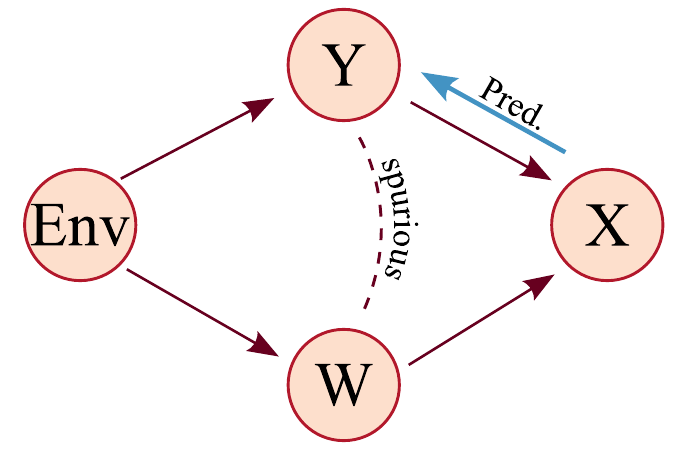}
    \caption{Reasoning about generalisation of a prediction task with a causal graph. Anti-causal prediction and a spurious association that may lead to shortcut learning are illustrated.}
    \label{fig:generalisation}
\end{figure}

Imagine a prediction problem where the goal is to learn $P(Y|X)$, with the causal graph illustrated in Fig.\ \ref{fig:generalisation}. We consider an environment variable $Env$ which controls the relationship between $Y$ and $W$. $Env$ is a confounder $Y \leftarrow Env \rightarrow W$ and $X$ is caused by the two variables $Y \rightarrow X \leftarrow W$.  

Firstly, we consider the view that most prediction problems are in the anti-causal direction \citep{Scholkopf2012anticausal, kilbertus2018generalization,Deml2021Anti,rosenfeld2021the}\footnote{We note that other seminal works \citep{Peters2016invariant,Arjovsky2019irm} consider prediction a causal task because prediction should copy a cognitive human process of generating labels given the data.}. That is, when making a prediction from a high-dimensional, unstructured variable $X$ (e.g.\ a brain image) one is usually interested in extracting and/or categorising one of its true generating factors $Y$ (e.g.\ gray matter volume). $P(X|Y)$, which represents the causal mechanism, $Y \rightarrow X$, is independent of $P(Y|Env)$, however $P(Y|X)$ is not as $P(Y|X) = P(X|Y)P(Y|env)/P(X)$. Thus $P(Y|X)$ changes as the environment changes.

Secondly, another (or many others) generating factor $W$ is often correlated with $Y$, which might cause the predictor to learn the relationship between $X$ and $W$ instead of the $P(Y|X)$. This is known as shortcut learning \citep{Geirhos2020} as it may be easier to learn the \textit{spurious correlation} than the required relationship. For example, suppose an imaging dataset $X$ is collected from two hospitals, $Env_1$ and $Env_2$. Hospital $Env_1$ has a large neurological disorder unit, hence a higher prevalence of AD status (denoted by $Y$), and uses a 3T MRI scanner (scanner type denoted by $W$). Hospital $Env_2$ with no specialist unit, hence a lower prevalence of AD, happens to use a more common 1.5T MRI scanner. The model will learn the spurious correlation between $W$ (scanner type) and $Y$ (AD status).

We can now describe several machine learning settings based on this causal perspective by comparing data availability at train and test time. Classical \emph{supervised learning} (or empirical risk minimisation (ERM) \citep{vapnik1999overview}) uses the strong assumption that the data from train and test sets are \emph{independent and identically distributed (i.i.d.)}, therefore we assign the same environment for both sets. \emph{Semi-supervised learning} \citep{chapelle2009semi} is a case where part of the training samples are not paired to annotations. \emph{Continual (or Lifelong) learning} considers the case where data from different environments are added after training, and the challenge is to learn new environments without forgetting what has initially been learned. In \textit{domain adaptation}, only unpaired data from the test environment is available during training. \textit{Domain generalisation} aims at learning how to become invariant to changes of environment, such that a new (unseen in training data) environment can be used for the test set. Enforcing \textit{fairness} is important when $W$ is a sensitive variable and the train set has $Y$ and $W$ spuriously\footnote{We  use the term \textit{spurious} for features that correlate but do not have a causal relationship between each other.} correlated due to a choice of environment. Finally, learning from \textit{imbalanced} datasets can be seen under this causal framework when a specific $Y = y$ have different numbers of samples because of the environment, but the test environment might contain the same bias towards a specific value of $Y$.

\section{Research Directions in Causal Machine Learning}
\label{sec:solutions}

Having discussed the utility of CML for healthcare including complex multimodal, temporal and unstructured data, the final section of this paper discusses some future research directions. We discuss CML according to the three categories defined in Section\ \ref{sec:intro}:
\begin{enumerate*}[label=(\roman*)]
    \item Causal Representation Learning;
    \item Causal Discovery; and
    \item Causal Reasoning.
\end{enumerate*}

\subsection{Causal Representations}
\label{subsec:causal_representation_learning}

Representation learning \citep{6472238} refers to a \textit{compositional} view of machine learning. Instead of a mapping between input and output domains, we consider an intermediate representation that captures concepts about the world. This notion is essential when considering learning and reasoning with real healthcare data. High-dimensional and unstructured data, as considered in Section\ \ref{sec:complex_data}  \ref{subsec:generalisation}, are not organised in units that can be directly used in current causal models. In most situations, the variable of interest is not, for instance, the image itself, but one of its generating factors, for instance gray matter volume in the AD example.  

\emph{Causal} representation learning \citep{scholkopf2021toward} extends the notion of learning factors about the world to modelling the relationships between variables with causal models. In other words, the goal is to model the representation domain $\mathcal{Z}$ as an SCM as in Section\ \ref{sec:whatscausality}  \ref{subsec:scms}. Causal representation learning builds on top of the \emph{disentangled} representation learning literature \citep{Higgins2017betaVAELB,chen2016infogan,liu2021tutorial} towards enforcing stronger inductive bias as opposed to assumptions of factor independence commonly pursued by disentangled representations. The idea is to reinforce a hierarchy of latent variables following the causal model, which in turn should follow the real data generation process.

\subsection{Causal Discovery}
\label{subsec:causal_discovery}

Performing RCTs is very expensive and sometimes unethical or even impossible. For instance, to understand the impact of smoking in lung cancer, it would be necessary to force random individuals to smoke or not smoke. Most real data are observational and discovering causal relationships between the variables is more challenging. Considering a setting where the causal variables are \textbf{known}, \textit{causal discovery} is the task of learning the direction of causal relationships between the variables. In some settings, we have many input variables and the goal is to construct the graph structure that best describes the data generation process. 

Extensive background has been developed over the last 3 decades around discovering causal structures from observational data, as described in recent reviews of the subject \citep{peters2017elements,Glymour2019frontiers,Nogueira2021,vowels2021dya}. Most methods rely on conditional independence tests, combinatorial exploration over possible DAGs and/or assumptions about the data generation process' function class and noise distribution ( e.g.\ the true causal relationships assumed to be linear, with additive noise or that the exogenous noise has a Gaussian distribution) for finding the causal relations of given causal variables. 

Causal discovery is still an open area of research and some of the major challenges in discovering causal effects \cite{peters2017elements, Prosperi2020} from observational data are the inability to
\begin{enumerate*}[label=(\roman*)]
    \item identify all potential sources of bias (unobserved confounders); 
    \item select an appropriate functional form for all variables (model misspecification); and
    \item model temporal causal relationships.
\end{enumerate*}

\subsection{Causal Reasoning}

It has been conjectured that humans internally build generative causal models for imagining approximate physical mechanisms through intuitive theories \citep{kilbertus2018generalization}. Similarly, the development of models that leverage the power of causal models around interventions would be useful. The causal models from Sections\ \ref{sec:whatscausality} \ref{subsec:scms} and \ref{subsec:potential_outcomes} can be formally manipulated for measuring the effects of interventions. Using causal models for quantifying the effect of interventions and pondering about the best decision is known as \textit{causal reasoning}.

Causal reasoning also refers to the ability to answer counterfactual queries about historical situations, such as``What would have happened if the patient had received alternative treatment X?''. We elaborated at length on the benefits of counterfactuals in the healthcare context in Section\ \ref{sec:causal_healthcare} \ref{subsec:precision_medicine}. 

One of the key benefits of reasoning causally about a problem domain is transparency, by offering a clear and precise language to communicate assumptions about the collected data \citep{chou2021counterfactuals,rudin2019stop,castro2020causality} as detailed in Section\ \ref{sec:causal_healthcare} \ref{subsec:data_gen_process}. In a similar vein, models whose architecture mirrors an assumed causal graph can be desirable in applications where interpretability is important \citep{moraffah2020causal}.

The main challenges in causal reasoning with ML relate to performing interventions with complex data representations and functions. Strategies for counterfactual prediction are simpler with scalar variables and linear functions. Interventions can have qualitatively distinct behaviours and should be understood as acting on high-level features rather than purely on the raw data. However, estimating counterfactuals in image features \citep{Pawlowski,Sanchez2022diffscm}, for example, requires invertible mechanisms such as normalising flows \citep{papamakarios2019normalizing} and/or methods for variational inference \citep{Kingma2014} which have their own complexities. Another open problem is how to deal with multimodal data e.g.\ images, text, age, sex and genetic data in a healthcare scenario as detailed in Section \ref{sec:complex_data} \ref{subsec:dimension_modal}.

\section{Conclusion}

We have described the importance of considering causal machine learning in healthcare systems. We highlighted the need to design systems that take into account the data generation process. A causal perspective on machine learning contributes to the goal of building systems that are not just performing better (e.g.\ achiever higher accuracy), but are able to reason about potential effects of interventions at population and individual levels, closing the gap towards realising precision medicine. 

We have discussed key pressing challenges in precision medicine and healthcare, namely, utilising multi-modal, high-dimensional and unstructured data to make decisions that are generalisable across environments and take into account temporal information. We finally proposed opportunities drawing inspiration from causal representation learning, causal discovery and causal reasoning towards addressing these challenges.

\bibliographystyle{plainnat}
\bibliography{references1}

\end{document}